\documentclass[letterpaper, 10 pt, conference]{ieeeconf}
\IEEEoverridecommandlockouts
\usepackage{amsmath,amsfonts}
\usepackage{algorithmic}
\usepackage{algorithm}
\usepackage{array}
\usepackage{textcomp}
\usepackage{stfloats}
\usepackage{url}
\usepackage{verbatim}
\usepackage{graphicx}
\usepackage{cite}
\usepackage{balance}
\usepackage{stfloats}
\usepackage{booktabs}
\usepackage{color}
\usepackage[dvipsnames]{xcolor}
\usepackage{dsfont}
\usepackage{colortbl}
\usepackage{bm}
\definecolor{DBlue}{HTML}{1E90FF}
\usepackage{soul}

\usepackage{tikz}
\usepackage{tikz-3dplot}
\usetikzlibrary{positioning, shapes.geometric, arrows.meta, fit, calc}
\usepackage{ragged2e}

\usepackage{makecell}
\usepackage{booktabs}
\usepackage{amssymb}
\usepackage{bbding}
\usepackage{pifont}
\usepackage{wasysym}
\usepackage{subfigure}

\definecolor{companionlight}{RGB}{240,245,255}
\definecolor{companiondark}{RGB}{220,235,255}
\definecolor{flightlight}{RGB}{240,255,240}
\definecolor{flightdark}{RGB}{220,255,220}
\definecolor{hardwarelight}{RGB}{255,250,240}
\definecolor{hardwaredark}{RGB}{255,240,220}
\definecolor{actuatorlight}{RGB}{255,240,240}
\definecolor{actuatordark}{RGB}{255,220,220}
\definecolor{spraylight}{RGB}{250,240,255}
\definecolor{spraydark}{RGB}{235,225,255}

\newcommand{\blue}[1]{\textcolor{black}{#1}}

\usepackage[urlcolor=DBlue, 
citecolor=DBlue, 
linkcolor=DBlue, 
colorlinks=true,]{hyperref}

\begin{document}

\title{\LARGE \bf 
Autonomous Close-Proximity Photovoltaic Panel Coating Using a Quadcopter}

\author{Dimitri Jacquemont, Carlo Bosio, Teaya Yang, Ruiqi Zhang, \"Ozg\"ur \"Or\"un, \\ Shuai Li, Reza Alam, Thomas M. Schutzius, Simo A. M\"akiharju,  and Mark W. Mueller \thanks{The authors are with High Performance Robotics Laboratory, Department of Mechanical Engineering, University of California Berkeley, CA 94720, United States.}%
}

\markboth{Journal of \LaTeX\ Class Files,~Vol.~14, No.~8, August~2021}%
{Shell \MakeLowercase{\textit{et al.}}: A Sample Article Using IEEEtran.cls for IEEE Journals}


\maketitle

\begin{abstract}
Photovoltaic (PV) panels are becoming increasingly widespread in the domain of renewable energy, and thus, small efficiency gains can have massive effects. Anti-reflective and self-cleaning coatings enhance panel performance but degrade over time, requiring periodic reapplication. Uncrewed Aerial Vehicles (UAVs) offer a flexible and autonomous way to apply protective coatings more often and at lower cost compared to traditional manual coating methods.
We present a quadcopter-based system, equipped with a liquid dispersion mechanism, designed to automate such tasks. The localization stack only uses onboard sensors, relying on visual-inertial odometry and the relative position of the PV panel detected with respect to the quadcopter. The control relies on a model-based controller that accounts for the ground effect and the mass decrease of the quadcopter during liquid dispersion. We validate the autonomy capabilities of our system through extensive indoor and outdoor experiments.
\end{abstract}


\section{Introduction}

Production of energy from photovoltaic (PV) panels is expected to become the largest source of renewable electricity by 2029 \cite{iea2024renewables}. The energy output of PV systems can be significantly reduced by the accumulation of organic matter and dust, leading to efficiency losses of over $50\%$ in some locations, the generation of hotspots, and degradation of panels \cite{aghaei2022review}. While automated cleaning systems provide effective dust removal, they require additional costs for installation, maintenance, and power consumption, and introduce added complexity and potential failure points to the system~\cite{FARROKHIDERAKHSHANDEH2021101518}. Self‑cleaning coatings offer a passive maintenance solution: they reduce soiling accumulation and lower maintenance requirements without additional operational energy input \cite{INGOLE2025106794, padhan2025high}. 

Photovoltaic panels also suffer from optical losses of about 4\% because light is refracted and reflected at the glass interface \cite{law2023performance}. To mitigate these losses, panels can be treated with anti‑reflective coatings. These coatings can be deposited by various methods, with spray coating widely adopted for its flexibility and simplicity \cite{INGOLE2025106794}. Several commercial products are available for PV applications, including Hydrasol \cite{hydrasol}, a hydrophobic coating, and TriNANO AR \cite{trinano}, which provides both self‑cleaning and anti‑reflective functions.

Despite efforts to incorporate UV‑blocking and wear‑resistant layers, these coatings gradually degrade under environmental exposure and erosion, diminishing their effectiveness over time \cite{aghaei2022review}. According to \cite{law2023performance} and \cite{https://doi.org/10.1002/pip.3575}, coating lifetimes are often limited to around 8 years or less, a fraction of the typical 25-year lifespan of PV modules, necessitating periodic reapplication to maintain performance. Given the vast scale of modern solar farms, re-coating is logistically challenging. To the best of our knowledge, no autonomous strategies have been developed to automate the reapplication process after they have degraded, either due to the absence of automated solutions or concerns about warranty violations.

While autonomous coating solutions are limited, some autonomous products exist for post-installation PV panel cleaning. For example, BladeRanger \cite{bladeranger} employs an autonomous quadcopter equipped with a long-reach end-effector for surface cleaning. This design minimizes liquid spray dispersion due to wind by maintaining the nozzle at a close distance from the panel surface, though small attitude errors can result in large deviations at the end-effector. In contrast, companies such as SolarCleano \cite{solarcleano} use ground-based robotic cleaners that move directly across panel surfaces. 
While fluid delivery using quadcopters has been explored in applications such as large-scale agricultural spraying \cite{5714}, research on high-precision, close-proximity fluid delivery remains limited. A quadcopter able to paint vertical surfaces was presented in \cite{8379422}, but it is used indoor and it does not account for ground effect or mass variation disturbances. In this work, we propose a solution filling this gap.

\begin{figure}[tb]
\centerline{\includegraphics[width=0.95 \columnwidth, trim=800 700.0 800 500.0, clip]{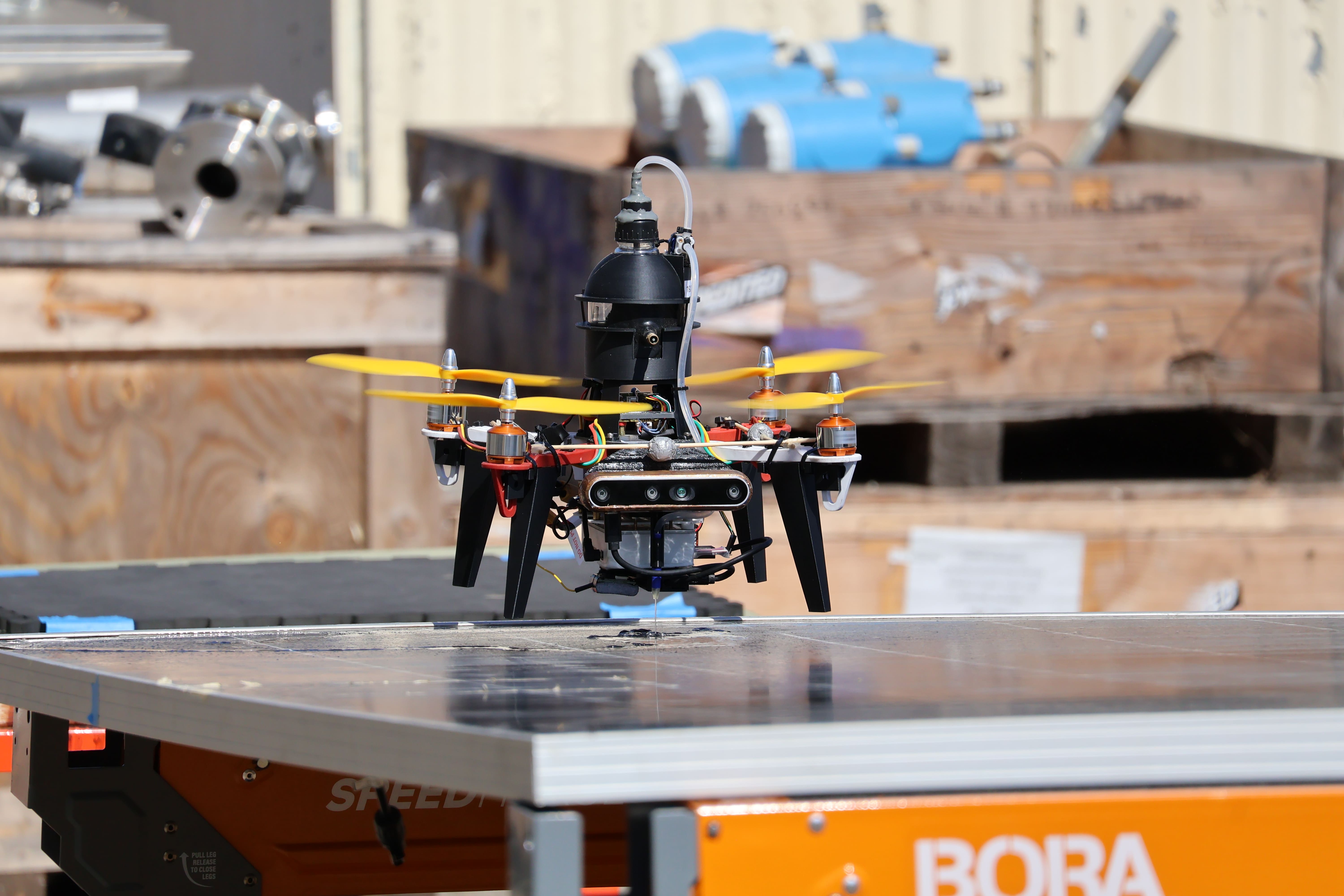}}
    \caption{The quadcopter flying over a $1.1 \times 2.3$~m photovoltaic panel, showing the reservoir mounted on top and the coating liquid being applied through a jet nozzle mounted at the bottom.}
\label{fig:drone_system}
\end{figure} 

An autonomous aerial system (Fig. \ref{fig:drone_system}) for reapplying coatings onto deployed PV panels offers several advantages over alternative approaches. Such systems can easily scale, contrary to ground-based or contact-based robots requiring infrastructure modifications. Ground-based robots struggle in challenging environments where PV panels are often installed, such as uneven terrain or rooftops, while quadcopters can cover vast solar farms without additional infrastructure requirements. However, developing such a system presents several technical challenges. Environmental disturbances such as wind, and notably complex wind fields generated around tilted panel surfaces, pose stability and control difficulties. Furthermore, a quadcopter must operate in close proximity to the panels to avoid unintentional dispersion of chemicals in the environment and to ensure uniform coating.

In this work, we present a fully integrated and scaled-down system for the autonomous in-field coating of PV panels. The platform combines a custom spraying mechanism with precise onboard state estimation using visual-inertial odometry (VIO). To enable robust low-altitude flight near panel surfaces despite aerodynamic disturbances, the control architecture incorporates a model-based thrust compensation for ground effect as well as the gradual mass reduction from liquid dispersion. We validate the system through outdoor experiments, demonstrating the coating deposition performances and low-altitude flight over PV arrays. This prototype aims at demonstrating the feasibility of coating individual PV panels, with the goal of inspiring future work toward large-scale solar farm applications.

The remainder of this paper is organized as follows. Section \ref{sec:robotsys} describes the robotic platform, including the hardware architecture and liquid dispersion system. Section \ref{sec:autonomy} outlines the autonomy stack and control system. Section \ref{sec:compensation} details the compensation methods for ground effect and mass variation disturbances. Section \ref{sec:expsetup} presents the experimental setup and results, covering VIO performance evaluation, comparative analysis of compensation strategies, and qualitative validation through outdoor testing.

\section{Hardware design \& trade-offs}
\label{sec:robotsys}

\begin{figure}[tb]
\centerline{\includegraphics[width=0.95 \columnwidth]{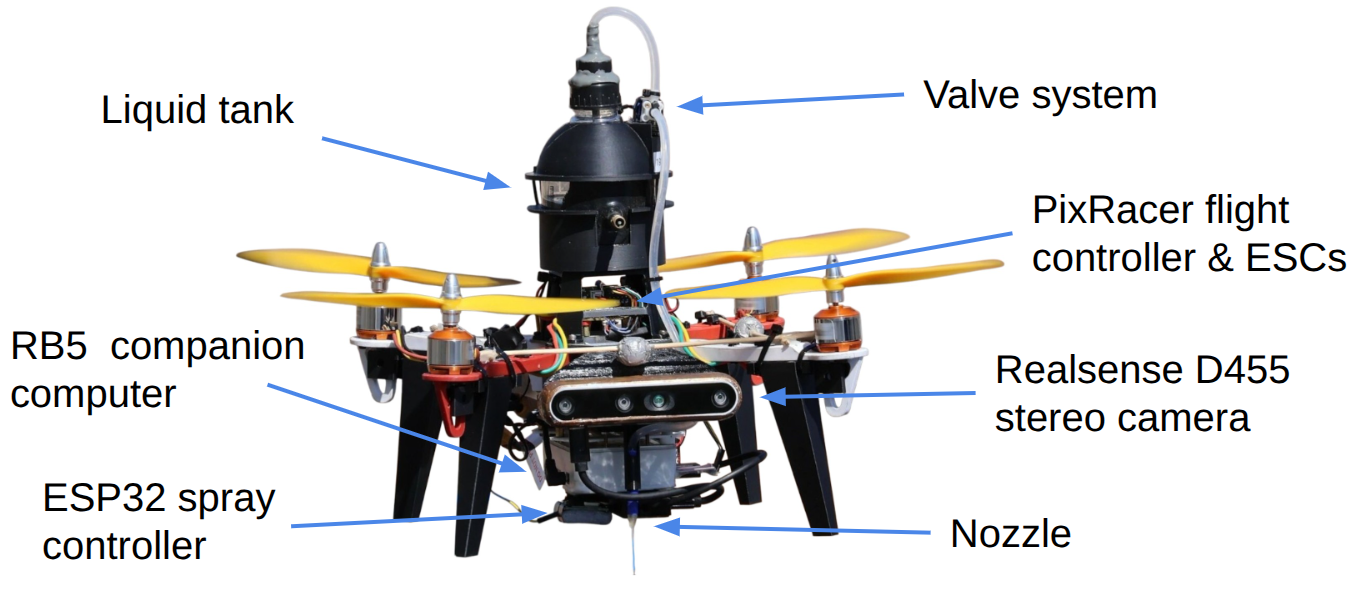}}
\caption{Quadcopter hardware configuration. The platform integrates a Qualcomm RB5 companion computer for high-level autonomy, a PixRacer R15 flight controller and ESCs for low-level control, an Intel RealSense D455 stereo camera for visual-inertial odometry (VIO), and an ESP32-based spray controller operating the valve of the dispersion system.}
\label{fig:hwsys}
\end{figure}

This section describes the hardware components and decisions in the overall UAV design (Fig. \ref{fig:hwsys}), and the details of the liquid dispersion system.

\subsection{Aerial Platform}

The quadcopter is designed to fit within a $30\,\mathrm{cm} \times 30\,\mathrm{cm} \times 30\,\mathrm{cm}$ volume for safe in-laboratory testing.
It has a diagonal motor-to-motor span of $34 \, \mathrm{cm}$ and a total take-off weight of $1.56 \, \mathrm{kg}$. This includes the $300\,\mathrm{g}$ liquid dispersion system filled with $150\,\mathrm{ml}$ of liquid. The system is powered by a 3-cell $5000~\rm{mAh}$ LiPo battery. At hover, the propulsion system operates at approximately $54\%$ of its maximum thrust capacity. The vehicle is equipped with a Qualcomm RB5 companion computer, which handles high-level autonomy logic, VIO processing, PV panel corner detection, and the control and compensation systems. Low-level logic is handled by a Pixracer R15 flight controller running a custom fork of the PX4 firmware \cite{px4}. An Intel RealSense D455 provides visual sensing, used together with the PixRacer's inertial measurement unit (IMU) to enable VIO. The sensors are calibrated together using the Kalibr toolbox \cite{6696514}.

\subsection{Liquid Dispersion System}
\label{subsec:liquid_dispersion_syst}

With the goal of coating a $1.1 \times 2.3~\rm{m}$ PV panel, the components are selected to ensure the system can carry at least $100 \, \mathrm{ml}$ of coating liquid of density approximately $1.0\,\mathrm{g\cdot cm^{-3}}$  while remaining lightweight and compact. The coating system consists of a $250~\rm{ml}$ pressurizable plastic container, filled with liquid to $60\%$ capacity ($150~\rm{ml}$). The container is manually pressurized up to $240~\rm{kPa}$ using a Schrader valve and an external pump, which provides consistent spray or jet performance. A solenoid valve regulates the flow of liquid to a nozzle and is digitally actuated via the onboard co-computer. The final design fits within a $20\,\mathrm{cm} \times 20\,\mathrm{cm} \times 30\,\mathrm{cm}$ volume and weighs approximately $150\,\mathrm{g}$ when empty.

For durability under pressure, all joints and fittings are sealed using epoxy and silicon adhesive. The solenoid valve is controlled via an ESP32 microcontroller, which receives commands from the RB5. Actuation is performed through a transistor circuit powered by a $5 \, \mathrm{V}$ DC-DC buck converter, which steps down the main $12 \, \mathrm{V}$ battery voltage. When triggered, the transistor allows current to flow in the solenoid valve, releasing the pressurized liquid through the nozzle.



\section{Software stack}
\label{sec:autonomy}

\begin{figure*}[tb]
\centering
\resizebox{0.9 \textwidth}{!}{
\begin{tikzpicture}[
    companion_block/.style={rectangle, draw, fill=companiondark, text width=3.5cm, text centered, minimum height=1.0cm, font=\small, align=center},
    flight_block/.style={rectangle, draw, fill=flightdark, text width=3cm, text centered, minimum height=1.0cm, font=\small, align=center},
    spray_block/.style={rectangle, draw, fill=spraydark, text width=3cm, text centered, minimum height=1.0cm, font=\small, align=center},
    hardware_block/.style={rectangle, draw, fill=hardwaredark, text width=2.2cm, text centered, minimum height=1.0cm, font=\small, align=center},
    actuator_block/.style={rectangle, draw, fill=actuatordark, text width=2.2cm, text centered, minimum height=1.0cm, font=\small, align=center},
    container/.style={rectangle, draw, thick, rounded corners=5pt},
    arrow/.style={-{Stealth[length=3mm]}, thick},
    elbow/.style={-{Stealth[length=3mm]}, thick, rounded corners=3pt},
    node distance=0.8cm and 1.5cm
]

\node[companion_block, opacity=0] (highlevel) {High Level Logic};
\node[companion_block, opacity=0, right=of highlevel] (trajectory) {Trajectory Generation};
\node[companion_block, opacity=0, above=0.2cm of trajectory] (masscomp) {Mass Compensation};
\node[companion_block, opacity=0, above=0.4cm of masscomp] (spraysys) {Dispersion System Interface};
\node[companion_block, opacity=0, below=0.2cm of trajectory] (groundeffect) {Ground Effect Compensation};
\node[companion_block, opacity=0, below=0.2cm of groundeffect] (vio) {VIO};
\node[companion_block, opacity=0, below=0.4cm of vio] (cornerdet) {Corner Detection};

\node[companion_block, opacity=0, right=of trajectory] (ctrl) {Controller};
\node[flight_block, opacity=0, right=of ctrl] (lll) {Motor Control};
\node[spray_block, opacity=0, above=of lll] (sc) {Valve Controller};

\node[hardware_block, opacity=0, right=of lll] (motors) {Motors};
\node[hardware_block, opacity=0, above=of motors] (valve) {Valve};
\node[hardware_block, opacity=0, below=of motors] (imu) {IMU};
\node[hardware_block, opacity=0, below=of imu] (camera) {Stereo Camera};

\node[container, fill=companionlight, fit=(highlevel) (spraysys) (masscomp) (trajectory) (groundeffect) (vio) (cornerdet) (ctrl), 
      inner xsep=4pt,
      inner sep=3pt,
      label={[anchor=south west, font=\footnotesize\bfseries, yshift=-2pt]north west:Qualcomm RB5 Companion Computer}] (companion_box) {};

\node[container, fill=flightlight, fit= (lll), 
      inner xsep=3pt,
      inner sep=4pt,
      label={[anchor=south west, font=\footnotesize\bfseries, yshift=-2pt]north west:PixRacer Flight Controller}] (flight_box) {};

\node[container, fill=spraylight, fit= (sc), 
      inner xsep=3pt,
      inner sep=4pt,
      label={[anchor=south west, font=\footnotesize\bfseries, yshift=-2pt]north west:ESP32 Valve Controller}] (spray_mcu_box) {};

\node[container, fill=actuatorlight, fit=(motors) (valve), 
      inner xsep=3pt,
      inner sep=4pt,
      label={[anchor=south west, font=\footnotesize\bfseries, yshift=-2pt]north west:Actuators}] (hardware_box) {};

\node[container, fill=hardwarelight, fit=(imu) (camera), 
      inner xsep=3pt,
      inner sep=4pt,
      label={[anchor=south west, font=\footnotesize\bfseries, yshift=-2pt]north west:Sensors}] (hardware_box) {};

\node[companion_block] at (highlevel) {Flight Stage Manager};
\node[companion_block] at (trajectory) {Trajectory Generation};
\node[companion_block] at (masscomp) {Mass Compensation};
\node[companion_block] at (spraysys) {Liquid Dispersion System Interface};
\node[companion_block] at (groundeffect) {Ground Effect Compensation};
\node[companion_block] at (vio) {Localization};
\node[companion_block] at (cornerdet) {Corner Detection};
\node[companion_block] at (ctrl) {Cascaded Tracking Controller};
\node[flight_block] at (lll) {Motor Speed Controller};
\node[spray_block] at (sc) {Valve Controller};
\node[actuator_block] at (motors) {Motors};
\node[actuator_block] at (valve) {Valve};
\node[hardware_block] at (imu) {IMU};
\node[hardware_block] at (camera) {Stereo Camera};

\draw[arrow] (highlevel.east) -- (trajectory.west);
\draw[elbow] (highlevel.east) -- ++(0.7,0) |- (spraysys);
\draw[elbow] (highlevel.east) -- ++(0.7,0) |- (masscomp);
\draw[elbow] (highlevel.east) -- ++(0.7,0) |- (groundeffect);
\draw[elbow] (vio.west) -| (highlevel.south);
\draw[elbow] (cornerdet.west) -| (highlevel.south);
\draw[arrow] (vio.south) -- (cornerdet.north);
\draw[arrow] (spraysys.south) -- (masscomp.north);
\draw[elbow] (spraysys.east) -- ++(1.5,0) |- (sc.west);
\draw[arrow] (trajectory.east) -- (ctrl.west);
\draw[elbow] (masscomp.east) -- ++(0.7,0) |- (ctrl);
\draw[elbow] (groundeffect.east) -- ++(0.7,0) |- (ctrl);
\draw[arrow] (ctrl.east) -- (lll.west);
\draw[arrow] (sc.east) -- (valve.west);
\draw[arrow] (lll.east) -- (motors.west);
\draw[elbow] (imu.west) -| (ctrl.south);
\draw[elbow] (imu.west) -- ++(-4,0) |- (vio.east);
\draw[elbow] (camera.west) -- ++(-4,0) |- (vio.east);
\draw[elbow] (camera.west) -- ++(-2.0,0) |- (cornerdet.east);

\end{tikzpicture}
}
\caption{\textbf{Autonomy stack block diagram} showing the system architecture across five main components: the RB5 companion computer handles high-level autonomy including flight stage, trajectory generation, localization, PV panel detection, and compensation algorithms; the PixRacer flight controller executes low-level control logic; the ESP32 spray controller controls the valve, and sensors and actuators represent the physical interfaces. Arrows indicate data flow and control commands between modules.}
\label{fig:autonomy_system}
\end{figure*}
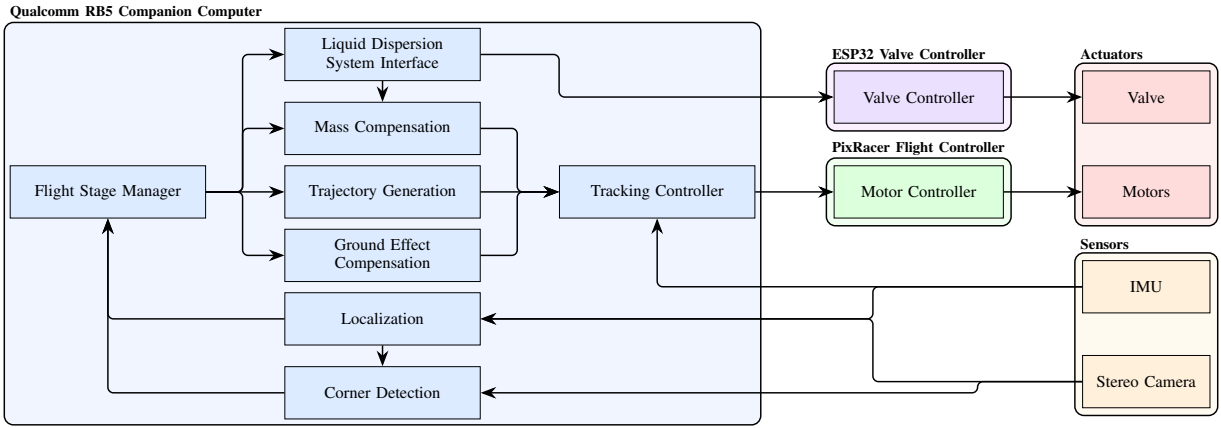

In this section, we present the proposed autonomy stack illustrated in Fig.~\ref{fig:autonomy_system}. The main autonomy components are discussed in the following paragraphs. 

\paragraph*{Localization}Contrary to real-time kinematics (RTK) GPS, which requires external infrastructure, visual-inertial odometry (VIO) provides accurate real-time state estimation without such requirement and can be used both indoors and outdoors.
Real-time state estimation is provided by a VIO module from the OpenVINS framework \cite{geneva2020openvins} operating at $500~\rm{Hz}$ using IMU and stereo camera data. OpenVINS is based on the multi-state constraint Kalman filter (MSCKF) \cite{4209642MSCKF}, consuming high-rate IMU data at $500~\rm{Hz}$ for state prediction and incorporating visual measurements from stereo cameras at $30~\rm{Hz}$. The estimator outputs a full 6-DoF pose along with linear and angular velocity estimates.

\paragraph*{Detection}The stereo camera data used for VIO also enables detection of PV panel corners relative to the quadcopter's position, with identified corner points fed to the trajectory generation module.
This is based on a pre-trained YOLO11n model \cite{khanam2024yolov11overviewkeyarchitectural} fine-tuned with a PV panel dataset \cite{realsolarpanelsoilshadingdataset}. YOLO11n is selected due to its efficiency in resource-constrained environments. The model runs inference at $2~\rm{Hz}$ on the RB5 companion computer's CPU, assuming that panel corners remain stationary throughout the flight, so continuous tracking after the initial estimation is unnecessary.



\begin{figure}[tb]
    \centering
    \subfigure[]{
        \includegraphics[width=0.45\columnwidth, trim=150 0.0 150 40.0, clip]{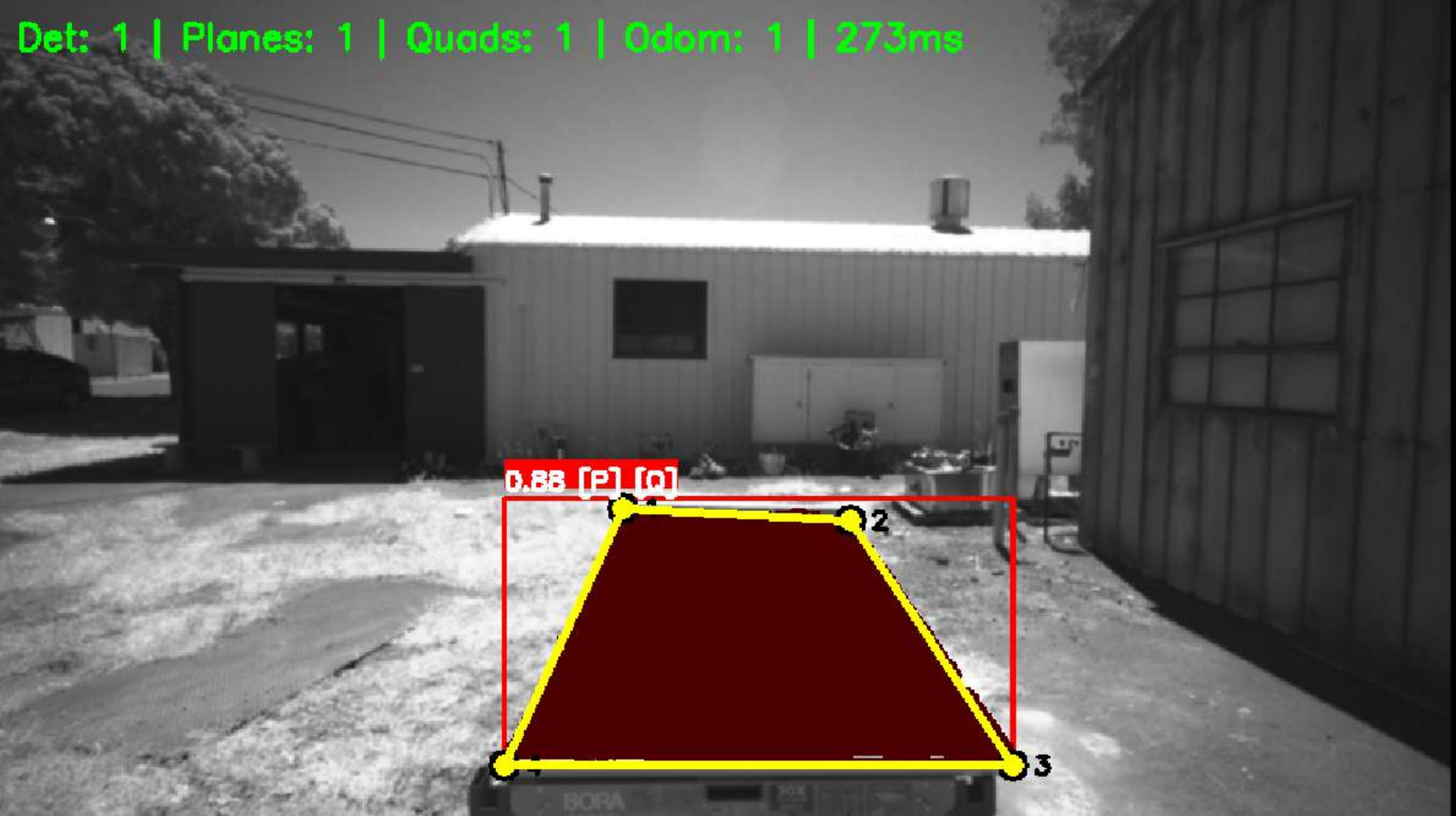}
        \label{fig:pv_seg_a}
    }%
    \hfill%
    \subfigure[]{
        \begin{tikzpicture}
            \node[anchor=south west,inner sep=0] (image) at (0,0) {
                \includegraphics[width=0.45\columnwidth]{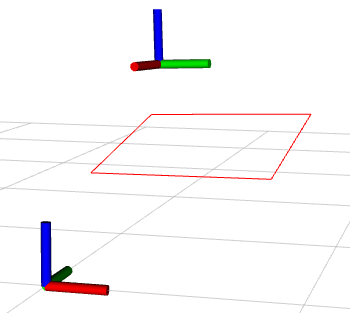}
            };
            \begin{scope}[x={(image.south east)},y={(image.north west)}]
                \node[text opacity=1, rounded corners=2pt] 
                    at (0.6,0.59) {\scriptsize Estimated};
                \node[text opacity=1, rounded corners=2pt] 
                    at (0.6,0.49) {\scriptsize PV panel};
                \node[text opacity=1, rounded corners=2pt] 
                    at (0.35,0.18) {\scriptsize World frame};
                \node[text opacity=1, rounded corners=2pt] 
                    at (0.65,0.87) {\scriptsize Drone frame};
            \end{scope}
        \end{tikzpicture}
        \label{fig:pv_seg_b}
    }
    \caption{\textbf{Photovoltaic panel detection pipeline}: (a) from image processing to world coordinate mapping. Panel detection with RANSAC plane fitting and corner extraction. (b) 3D visualization of relevant coordinate systems and detected PV panel. The panel coordinates are transformed from the drone frame (aligned with the IMU) to the world frame (i.e. from the VIO initialization).}
    \label{fig:pv_seg}
    \vspace{-0.4cm}
\end{figure}

The detection pipeline processes synchronized stereo camera images, depth data, and quadcopter state estimates. For inference, grayscale images downsampled to $320 \times 320$ pixels are used, retaining only detections in the lower image portion where PV panels are expected. For each detection, depth pixels within the bounding box form a 3D point cloud through which a plane is fitted using RANSAC \cite{ransac}, generating a binary mask of plane-belonging pixels (Fig.~\ref{fig:pv_seg_a}). Corner extraction identifies four extreme points from the main connected component within the plane mask. These 3D corners are then transformed to world coordinates using the estimated quadcopter pose (Fig.~\ref{fig:pv_seg_b}). A key limitation of this approach is the dependence on the 3D point cloud quality, which can degrade significantly due to reflections on the panel surface. \blue{In case the vision pipeline fails at detecting the solar panel, the system aborts the coating task and lands.}



\paragraph*{Control}A high level trajectory generation module provides a simple coverage trajectory based on the corner positions of the PV panel. Some parameters, such as reference flight speed and spacing between consecutive sweep passes, allow to generate diverse sweeping trajectories.

The controller employs a cascaded three-stage architecture: position control generates attitude and thrust commands from waypoints, attitude control produces rate commands, and rate control converts these commands into motor torques. The active panel detection approach allows the control stack to adapt to any panel slope by generating waypoints at a fixed distance from the fitted panel plane.

To compensate for complex aerodynamic effects around PV panels and payload mass variation, several strategies were considered. Model-based control was preferred over learned policies, as it provides stronger guarantees for stability and reliability in safety-critical, close-proximity flight. The adopted approach incorporates a ground effect model and ejected mass estimation to mitigate ground-effect disturbances and position errors due to payload mass variation. An integrator term in the position controller compensates for the slowly varying battery voltage changes throughout flight.

\section{Disturbance Compensation}
\label{sec:compensation}

To improve quadcopter trajectory tracking during liquid dispensing operations above PV panels, two compensation methods are employed. The first compensates for ground effect disturbances caused by the quadcopter's close proximity to the panel surface, while the second compensates for flight dynamics changes caused by decreasing payload mass as liquid is dispensed.

\paragraph*{Quadrotor Dynamics}

\begin{figure}[tb]
\centering
\resizebox{0.7\columnwidth}{!}{
\tdplotsetmaincoords{60}{120}
\begin{tikzpicture}[tdplot_main_coords, scale=1.5]
\coordinate (O) at (0,0,0);
\draw[thick, black] (-1.5,0,0) -- (1.5,0,0);
\draw[thick, black] (0,-1.5,0) -- (0,1.5,0);
\draw[thick] (-1.5,0,0) circle (0.3);
\draw[thick] (1.5,0,0) circle (0.3);
\draw[thick] (0,-1.5,0) circle (0.3);
\draw[thick] (0,1.5,0) circle (0.3);
\draw[thick, ->] (O) -- ({0.7*cos(45    )},{0.7*sin(45)},0) node[anchor=south west]{$x_b$};
\draw[thick, ->] (O) -- ({0.7*cos(135)},{0.7*sin(135)},0) node[anchor=south east]{$y_b$};
\draw[thick, ->] (O) -- (0,0,0.7) node[anchor=south]{$z_b$};
\draw[thick, blue!70!black, ->] (-1.5,0,0) -- (-1.5,0,1.0) node[anchor=south] {$F_{2}$};
\draw[thick, blue!70!black, ->] (1.5,0,0) -- (1.5,0,1.0) node[anchor=south] {$F_{0}$};
\draw[thick, blue!70!black, ->] (0,-1.5,0) -- (0,-1.5,1.0) node[anchor=south] {$F_{1}$};
\draw[thick, blue!70!black, ->] (0,1.5,0) -- (0,1.5,1.0) node[anchor=south] {$F_{3}$};
\draw[thick, blue!70!black, <-] (-1.5,0,0) ++(0,0.4,0) arc (90:-90:0.4);
\node[anchor=center, blue!70!black] at (-1.5,-0.4,0.2) {$\omega_{2}$};
\draw[thick, blue!70!black, ->] (1.5,0,0) ++(0,0.4,0) arc (90:270:0.4);
\node[anchor=center, blue!70!black] at (1.5,0,-0.4) {$\omega_{0}$};
\draw[thick, blue!70!black, ->] (0,-1.5,0) ++(0.4,0,0) arc (0:-180:0.4);
\node[anchor=center, blue!70!black] at (0,-1.5,-0.4) {$\omega_{1}$};
\draw[thick, blue!70!black, <-] (0,1.5,0) ++(0.4,0,0) arc (0:180:0.4);
\node[anchor=center, blue!70!black] at (0,1.5,-0.4) {$\omega_{3}$};
\draw[thick, red!70!black, ->] (O) -- (0,0,-1.2) node[anchor=north]{$mg$};
\fill[black] (O) circle (0.07);
\coordinate (W) at (1.5,-1.5,-0.3);
\draw[thick, black, ->] (W) -- ++(0.5,0,0) node[anchor=north east]{$x_w$};
\draw[thick, black, ->] (W) -- ++(0,0.5,0) node[anchor=north west]{$y_w$};
\draw[thick, black, ->] (W) -- ++(0,0,0.5) node[anchor=south]{$z_w$};
\end{tikzpicture}
}
\caption{Quadcopter free body diagram.}
\label{fig:quadcopter_fbd}
\end{figure}
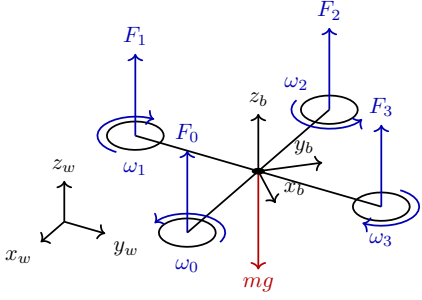

The quadcopter dynamics are described by the translational and rotational equations of motion \cite{mueller2025dynamics}. With reference to Fig. \ref{fig:quadcopter_fbd}, each rotor generates thrust $F_i = k_f \omega_i^2$ and torque $\tau_i = k_m \omega_i^2$, where $k_f$ and $k_m$ are respectively the torque and the force coefficient, and $\omega_i$ is the angular speed of the $i$-th propeller.
The total control inputs are:
\begin{align}
T &= k_f(\omega_{0}^2 + \omega_{1}^2 + \omega_{2}^2 + \omega_{3}^2) \label{eq:total_thrust},\\
\boldsymbol{\tau} &= \begin{bmatrix} k_f \frac{\ell}{\sqrt{2}} (-\omega_{0}^2 - \omega_{1}^2 + \omega_{2}^2 + \omega_{3}^2) \\ k_f \frac{\ell}{\sqrt{2}} (-\omega_{0}^2 + \omega_{1}^2 + \omega_{2}^2 - \omega_{3}^2) \\ k_m(\omega_{0}^2 - \omega_{1}^2 + \omega_{2}^2 - \omega_{3}^2) \end{bmatrix}, \label{eq:moments}
\end{align}
where $\ell$ is the vehicle's arm length. The translational dynamics in the world frame  are:
\begin{equation}
m\ddot{\boldsymbol{s}} = \begin{bmatrix} 0, \ 0, \ -mg \end{bmatrix}^T + \mathbf{R}_{bw} \begin{bmatrix} 0, \ 0, \ T \end{bmatrix}^T
\label{eq:translational_dynamics}
\end{equation}
where $m$ is the mass of the body, $g=9.81\,\mathrm{m\cdot s^{-2}}$ is the gravity acceleration, $\ddot{\boldsymbol{s}}$ is the acceleration of the body in the world frame, and $\mathbf{R}_{bw}$ is the rotation matrix that rotates vectors from the body frame to the world frame.
The rotational dynamics in the body frame are:
\begin{equation}
\mathbf{J}\dot{\boldsymbol{\omega}} + \boldsymbol{\omega} \times \mathbf{J}\boldsymbol{\omega} = \boldsymbol{\tau}
\label{eq:rotational_dynamics}
\end{equation}
where $\mathbf{J}$ is the inertia matrix, and $\boldsymbol{\omega}$ are the body angular rates.

\subsection{Ground Effect Compensation}
Ground effect compensation is implemented using an established model from \cite{cheeseman}, adapted for quadcopters in \cite{7260521autolanding}. 
The propeller thrust due to ground effect is given by:
\begin{align}
\frac{T_{\text{in}}}{T_{\text{out}}} = 1-\rho\left(\frac{r}{4\cdot h}\right)^{2}
\label{IGE}
\end{align}
where $T_{\text{in}}$ is the commanded input thrust, $T_{\text{out}}$ is the actual thrust generated by a propeller subject to ground effect, $\rho$ is a coefficient determined experimentally (e.g. in \cite{7260521autolanding}), $r$ is the propeller radius ($r=10~\rm{cm}$  in our case), and $h$ is the vercial distance of the propeller from the surface.
To determine $\rho$ experimentally, we hover at different heights above the ground and record both $h$ and the corresponding $T_{\text{in}}$ required to maintain stable hover. At hover, the actual thrust satisfies $T_{\text{out}} = mg$. Substituting this into Eq.~\eqref{IGE}, we use a least squares approximation over the collected data to estimate the optimal value of $\rho$ for our system ($\rho= 5.71$ ).

\begin{figure}[tb]
    \centering
    \resizebox{0.5\columnwidth}{!}{
    \begin{tikzpicture}[scale=1.2]
        \draw[thick, fill=gray!30, rotate=15] 
            (1,0) rectangle (5,3);
        
        \node[rotate=15] at (3,1.5) {\textbf{PLANAR SURFACE}};
        
        \coordinate (center) at (0.1,2);
        
        \draw[thick] (center) ++(-0.8,-0.8) -- ++(1.6,1.6);
        \draw[thick] (center) ++(-0.8,0.8) -- ++(1.6,-1.6);
        
        \draw[thick, fill=NavyBlue!20] (center) ++(-0.8,0.8) circle (0.4);
        \node at ([shift={(-0.8,0.8)}]center) {\small 0.0};
        
        \draw[thick, fill=red!20] (center) ++(0.8,0.8) circle (0.4);
        \node at ([shift={(0.8,0.8)}]center) {\small 1.0};
        
        \draw[thick, fill=NavyBlue!20] (center) ++(-0.8,-0.8) circle (0.4);
        \node at ([shift={(-0.8,-0.8)}]center) {\small 0.0};
        
        \draw[thick, fill=red!15] (center) ++(0.8,-0.8) circle (0.4);
        \node at ([shift={(0.8,-0.8)}]center) {\small 0.78};
        
        \fill (center) circle (0.05);
        
    \end{tikzpicture}%
    }
    \caption{Representation of the propeller rotational area $\alpha_i$ in the computation of in-ground-effect adapted motor force used.}
    \label{fig:percentage_area}
\end{figure}
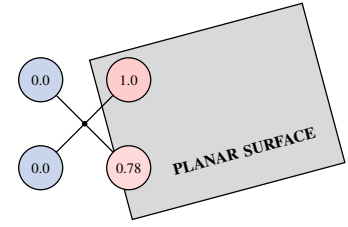

This model is applied individually to each rotor, which allows for a more accurate representation of disturbance torque and forces when entering a ground effect area from the side, or hovering on tilted surfaces. To capture the variation in force when flying close to a surface boundary, we calculate the fraction $\alpha_i \in \left[0, 1\right] \subset \mathbb{R}$ of each propeller’s rotational area overlapping with the surface below and scale the ground effect compensated force accordingly (Fig. \ref{fig:percentage_area}). At runtime, we compute the $\alpha_i$ from the onboard pose estimate and the panel estimate as in Fig. \ref{fig:pv_seg_a}.

\begin{figure}[tb]
\centerline{\includegraphics[width=1.0 \columnwidth]{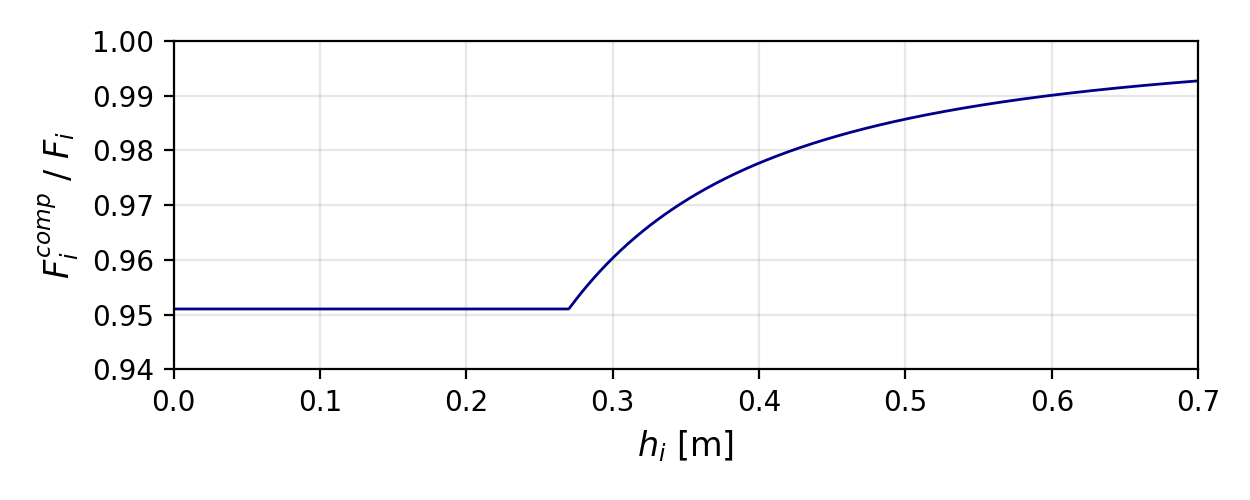}} 
\caption{Plot of the ratio $\mathbf{F}^\text{comp}_{i}/\mathbf{F}_{i}$, with $\alpha_i =1$ and $h_i^\text{des}=0.27\,\mathrm{m}$.}
\label{fig:ge_comp}
\end{figure}

Let $\mathbf{F}_{i}$ be the setpoint force for motor $i$, $i \in \{0, 1, 2, 3\}$, the compensated motor force in presence of ground effect $\mathbf{F}^\text{comp}_{i}$ can then be deduced and applied on each individual motor (Fig.~\ref{fig:ge_comp}).
\begin{align}
\frac{\mathbf{F}^\text{comp}_{i}}{\mathbf{F}_{i}} = 1- \alpha_i \rho\left(\frac{r}{4\cdot \max(h_{i}, h^\text{des}_{i})}\right)^{2}
\end{align} 
To prevent over-compensation when a rotor is below its desired altitude, the relative height in the ground effect calculation is clipped to $\max(h_{i}, h_i^\text{des})$, where $h_i^\text{des}$ and $h_i$ are the desired and actual heights above the surface underneath rotor $i$. This clipping ensures compensation remains constant when $h_{i} < h_i^\text{des}$, preventing excessive thrust reduction that could drive the quadcopter further below its commanded altitude. As the quadcopter approaches the surface, small height estimation errors can cause large thrust reduction errors, making this constraint critical to preventing dangerous compensation errors near the surface.

\subsection{Varying Mass Compensation}

As the coating liquid is offloaded from the quadcopter, the total mass of the system decreases and needs to be accounted for in the thrust required for it to hover. In the absence of this compensation, the distance quadcopter-panel increases throughout the coating process.

The approximate flow rate $f$ is determined experimentally by measuring the liquid volume ejected through the nozzle over a defined time interval at an initial container pressure of $240~\rm{kPa}$. When the valve is open and the liquid is being ejected, the the total mass estimate is updated as
\begin{equation}
    m = m_0 - \rho f \, \Delta t,
\end{equation}
where $m_0$ is the initial total mass, $\rho$ is the fluid density, $f$ is the experimentally measured flow rate, $\Delta t$ is the cumulative time the valve was open.

\section{Experimental Results}
\label{sec:expsetup}

We evaluate our system through indoor controlled testing and outdoor deployment. Each disturbance compensation method is tested individually under controlled conditions to assess performance. Qualitative system performance is evaluated through outdoor flight tests where the quadcopter is tasked to coat a full-scale PV panel surface.

\subsection{VIO Performance}

Accurate localization is essential for enabling precise panel coverage. To assess localization performances, the quadcopter executes a $12$-minute square trajectory in the laboratory using onboard VIO estimation, while motion capture data 
($200\,\mathrm{Hz}$, millimeter-level accuracy) serves as ground truth. The VIO system achieves a 3D position RMSE of $4.4 \, \mathrm{cm}$ and attitude RMSE of $2.6$ degrees, as presented in Table~\ref{tab:vio_results}. The magnitude of these localization errors is less than the ones introduced by external disturbances, such as ground effect and wind.

\begin{table}[ht]
    \centering
    \caption{3D performance of VIO when compared with motion capture.}
    \begin{tabular}{lc}
        \toprule
        \textbf{Variable} & \textbf{RMSE} \\
        \midrule
        Position [cm] & 4.4\\
        Attitude [deg] & 2.6\\
        \bottomrule
    \end{tabular}
    \label{tab:vio_results}
\end{table}

\subsection{Ground Effect Compensation}

\begin{figure}[tb]
\centerline{\includegraphics[width= 0.9 \columnwidth, trim=0 0.0 0 0, clip]{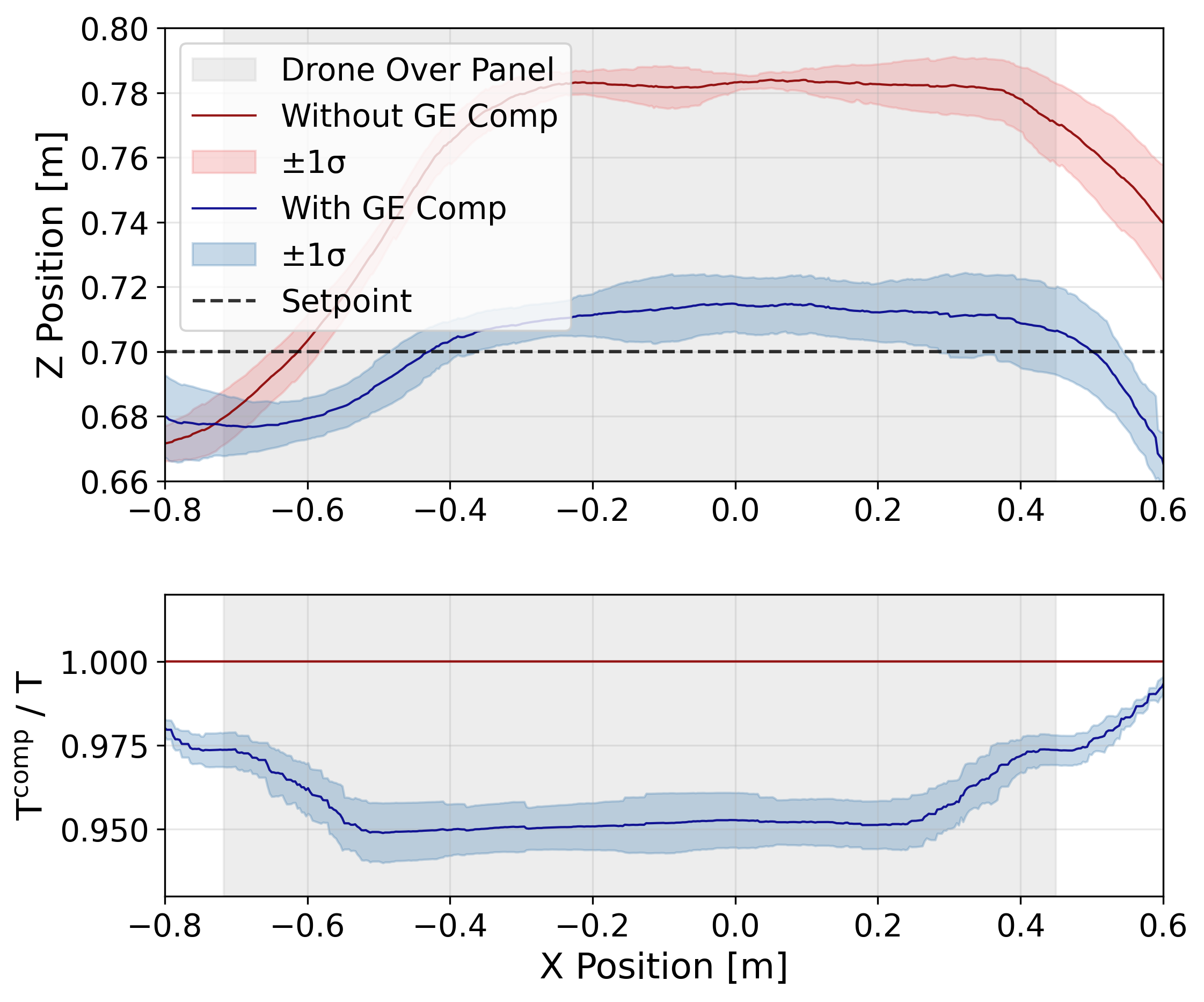}}
\caption{Example straight line trajectory $25\,\mathrm{cm}$ above a flat surface comparing ground effect compensation (GE comp) on/off. The compensation improves trajectory tracking and reduces total thrust by 5\% to account for ground effect disturbances, as seen on the second plot. The mean line and standard deviation bands are based on three runs.}
\label{fig:ge_compensation}
\end{figure}

To evaluate ground effect disturbances, the quadcopter is commanded to fly a straight trajectory along the $x$-axis at a height of $25~\rm{cm}$ above a flat surface. Such example height is a reference height that was empirically found effective for the coating task at hand. Tests are performed both with and without ground effect compensation, using motion capture for localization, and with the integrator term and mass compensation disabled. The results are presented in Fig.~\ref{fig:ge_compensation} and in Table~\ref{tab:ge_results}. Without compensation, the quadcopter exhibits an RMSE of $7.2~\rm{cm}$ when flying over the panel. With ground effect compensation enabled, the quadcopter achieves a significantly improved RMSE of $1.5~\rm{cm}$, validating the ground effect compensation approach.

\begin{table}[ht]
    \centering
    \caption{Altitude performance of the quadcopter flying in a straight trajectory with ground effect compensation on/off.}
    \begin{tabular}{lc}
        \toprule
        \textbf{Ground effect compensation} & \textbf{RMSE [cm]} \\
        \midrule
        Activated & \textbf{1.5}\\
        Deactivated & 7.2\\
        \bottomrule
    \end{tabular}

    \label{tab:ge_results}
\end{table}

To evaluate ground effect performance over a tilted surface, the drone was tasked to hover 25~$\mathrm{cm}$ above a surface with a 12.4° tilt angle, with the surface height increasing along the positive $y$-axis. The hovering test was conducted for 1 minute, both with and without compensation, focusing on Y and Z position errors. Table~\ref{tab:ge_results_2} shows that compensation activation reduced the RMSE along the $z$-axis from 11.9~$\mathrm{cm}$ to 1.8~$\mathrm{cm}$. The RMSE along the $y$-axis remained comparable between conditions, both Y and Z standard deviations decreased with compensation active, indicating more stable positioning.

\begin{table}[ht]

    \centering










    \caption{Performance evaluation of the drone hovering 27~cm above a $12.4^\circ$ tilted surface with ground effect compensation on/off.}
    \begin{tabular}{lcc}
        \toprule
        \textbf{Ground effect compensation} & \multicolumn{2}{c}{\textbf{RMSE [cm]}}\\
        \cmidrule(lr){2-3}
        & \textbf{Y} & \textbf{Z}\\
        \midrule
        Activated & 8.3 & \textbf{1.8}\\
        Deactivated & \textbf{8.1} & 11.9\\
        \bottomrule
    \end{tabular}
    \label{tab:ge_results_2}
\end{table}

\subsection{Mass Variation Compensation}

\begin{figure}[tb]
\centerline{\includegraphics[width=0.90\columnwidth,]{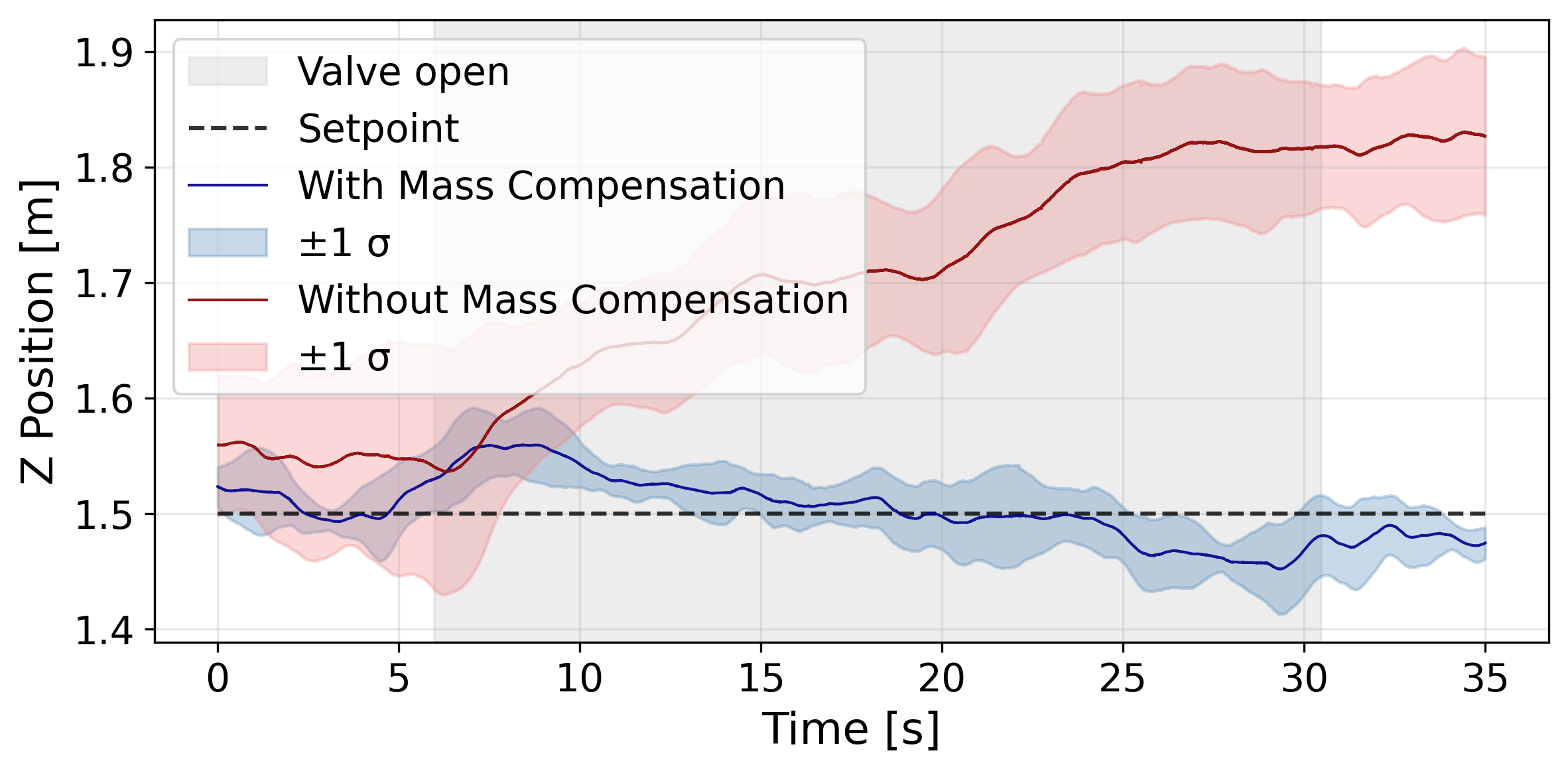}}
\caption{Hovering performance during water release at $240\,\mathrm{kPa}$ with mass compensation on/off. Compensation maintains steady altitude while uncompensated flight shows increasing altitude error as water is ejected. The mean line and standard deviation bands are based on three runs.}
\label{fig:mass_compensation}
\end{figure}

To evaluate the mass variation compensation system, the quadcopter is commanded to hover at a fixed position while liquid is released. Trials are performed both with and without mass compensation, using motion capture for localization, and with the integrator term and ground effect compensation disabled. Fig.~\ref{fig:mass_compensation} shows the quadcopter altitude over time. Without compensation, the quadcopter exhibits a steadily increasing altitude error as mass decreases, resulting in a final mean absolute altitude error of $32.7~\rm{cm}$. With compensation enabled, the quadcopter maintains stable hover throughout the discharge, with a reduced final mean altitude error of $2.7~\rm{cm}$, confirming the effectiveness of the approach.  
\subsection{Outdoor testing}
\begin{figure}[tb]
\centering
\subfigure[$xyz$-axis time-series of outdoor coverage flight with ground effect and mass variation compensation. The panel is sloped in the $yz$ plane. With reference to the $z$ estimate plot, at $t=0$ the quadcopter is flying over the lower edge of the panel at $z\simeq0.6\,\mathrm{m}$, and at $t=200$ it is flying over the higher edge at $z\simeq0.75\,\mathrm{m}$.]{
    \includegraphics[width=0.9\columnwidth]{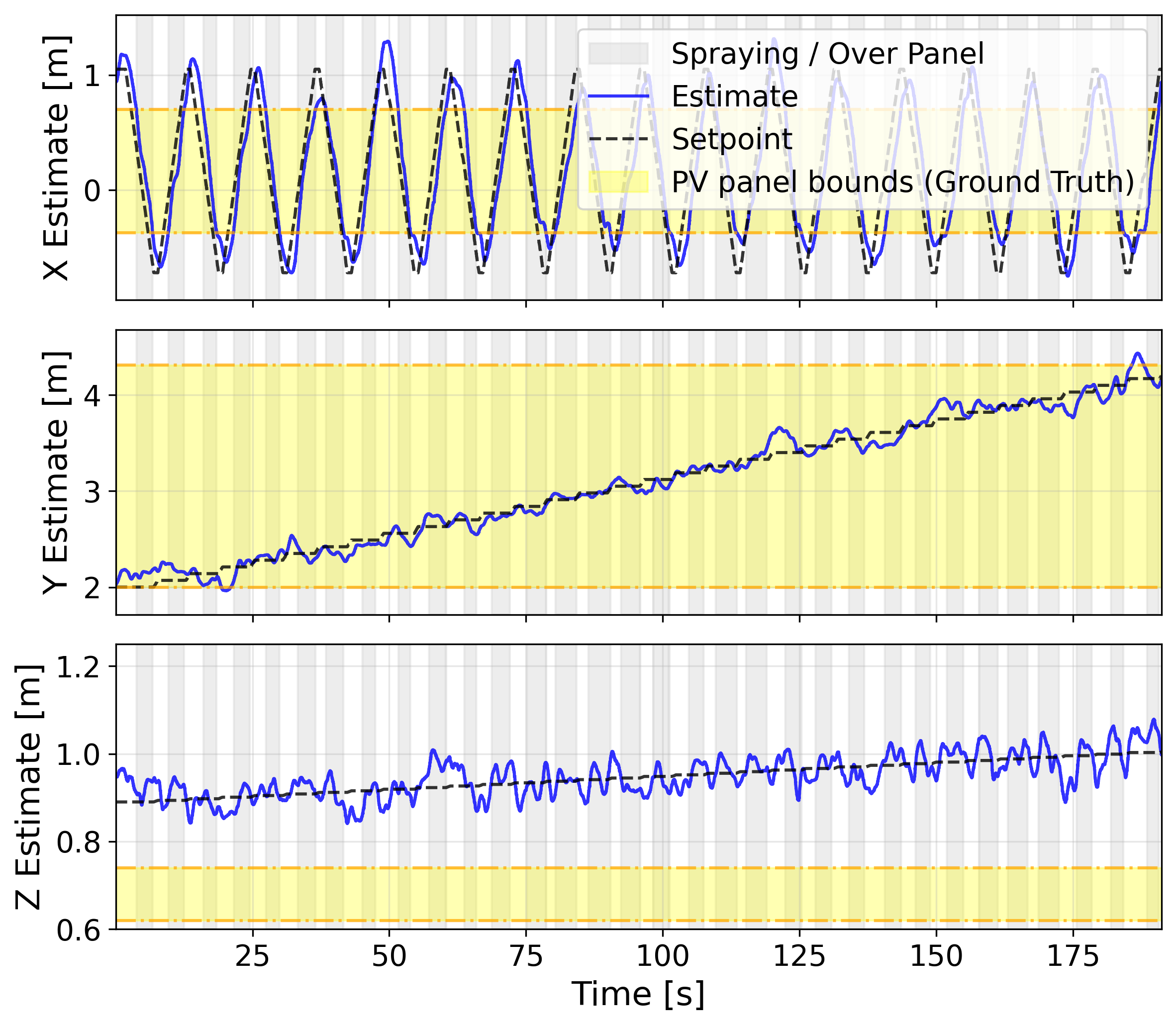}
    \label{fig:timeseries_wcomp}
}\\[1em]
\subfigure[$z$-axis time-series of outdoor coverage flight without ground effect and mass variation compensation.]{
    \includegraphics[width=0.9\columnwidth, trim=0 0.0 0 300.0, clip]{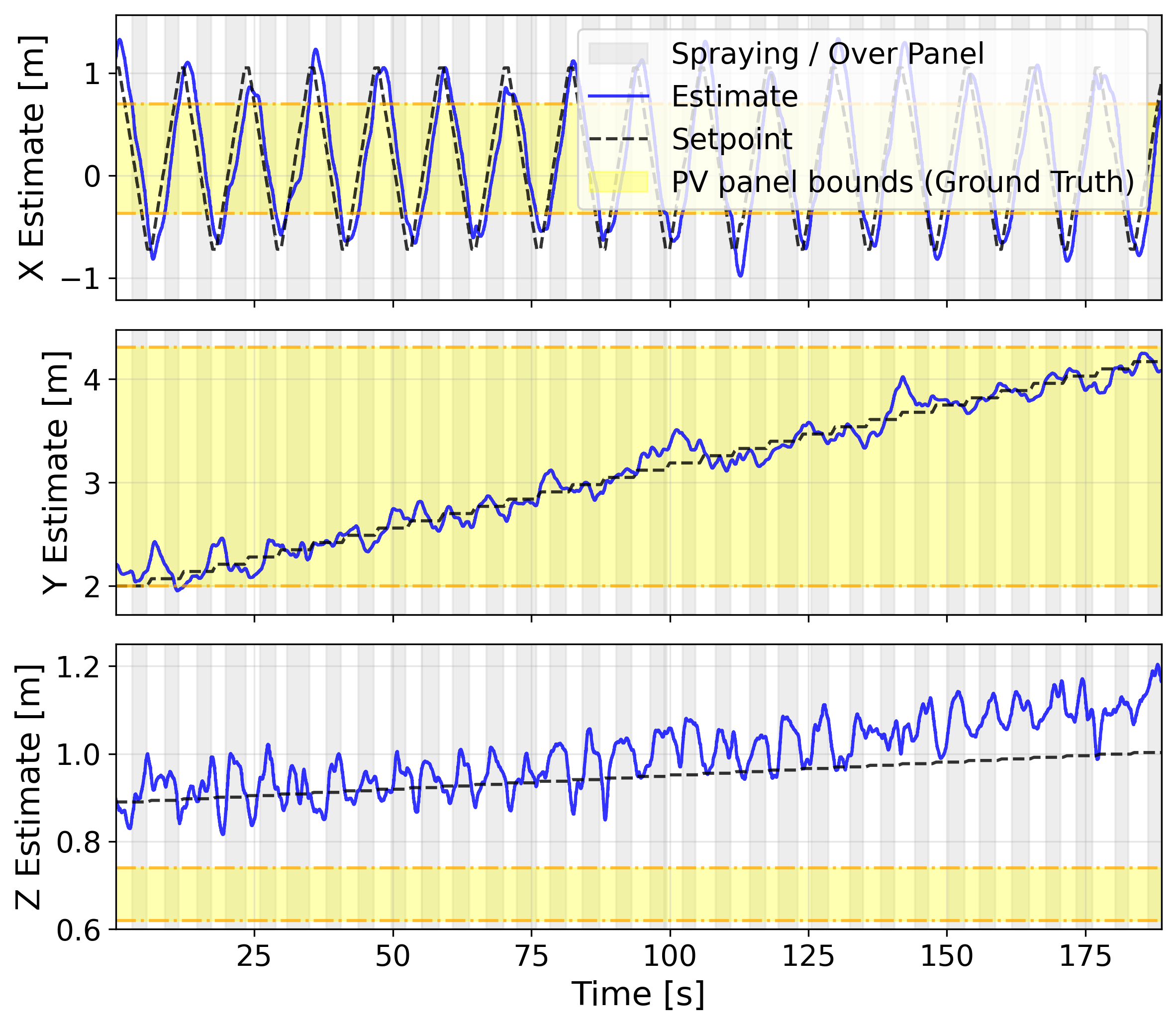}
    \label{fig:timeseries_wocomp}
}
\caption{Coverage flight over a $1.1 \times 2.3$~m photovoltaic panel with compensations on/off. The $x$ and $y$ directions are aligned with the short  and long sides of the PV panel, respectively. The sweeps are done in the $x$ direction. The $y$ coordinate is monotonically increased to avoid recoating of previously coated areas. The $z$ coordinate is monotonically increased because of the PV panel slope.}
\label{fig:coverage_traj_timeseries}
\end{figure}

The system was tested outdoors over a $1.1 \times 2.3~\rm{m}$ PV panel with $7~\rm{cm}$ inter-sweep distance using water as the test liquid. Two experimental runs evaluated the quadcopter behavior with compensations enabled 
(Fig.~\ref{fig:timeseries_wcomp}) and disabled (Fig.~\ref{fig:timeseries_wocomp}). Since the compensations primarily affect altitude control, only $z$-axis data is shown for the uncompensated run. Without compensation, the quadcopter exhibits increasing altitude offset over time due to mass reduction as liquid is dispensed through the nozzle, with significant tracking errors when navigating over the panel. With compensation active, $z$-axis errors remain minimal, enabling the quadcopter to maintain close proximity to the PV panel throughout the dispensing process. The tracking errors are mainly due to wind disturbances.

\begin{table}[ht]
    \centering
    \caption{Outdoor performance of the coverage flight over the panel, with compensations on/off.}
    \begin{tabular}{lccc}
        \toprule
         \textbf{Compensations} & \multicolumn{3}{c}{\textbf{RMSE [cm]}} \\
        \cmidrule(lr){2-4} 
        & \textbf{x} & \textbf{y} & \textbf{z} \\
        \midrule
        Activated & \textbf{29.0} & \textbf{7.6} & \textbf{3.0} \\
        Deactivated & 31.1 & 9.1 & 8.9 \\
        \bottomrule
    \end{tabular}
    \label{tab:outdoor_comp_results}
\end{table}

\begin{figure}[tb]
    \centering
    \subfigure[]{
        \setlength{\fboxsep}{0pt}%
        \setlength{\fboxrule}{0.1pt}%
        \fbox{\includegraphics[width=0.4\textwidth,angle=90]{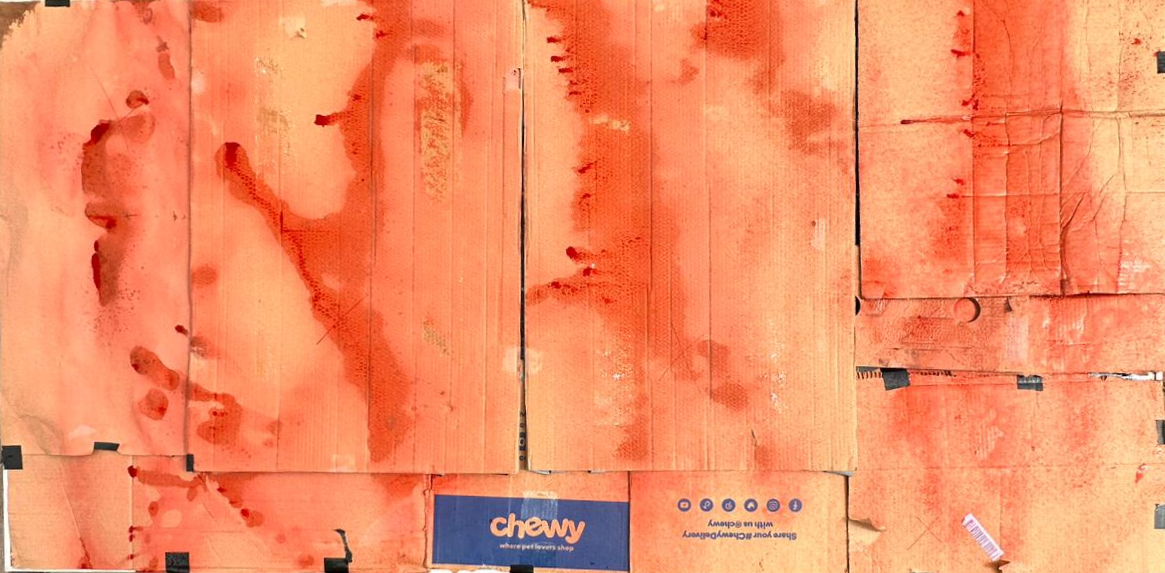}}
        \label{fig:liq_disp_a}
    }%
    \hfill%
    \subfigure[]{
        \setlength{\fboxsep}{0pt}%
        \setlength{\fboxrule}{0.1pt}%
        \fbox{\includegraphics[width=0.4\textwidth,angle=90]{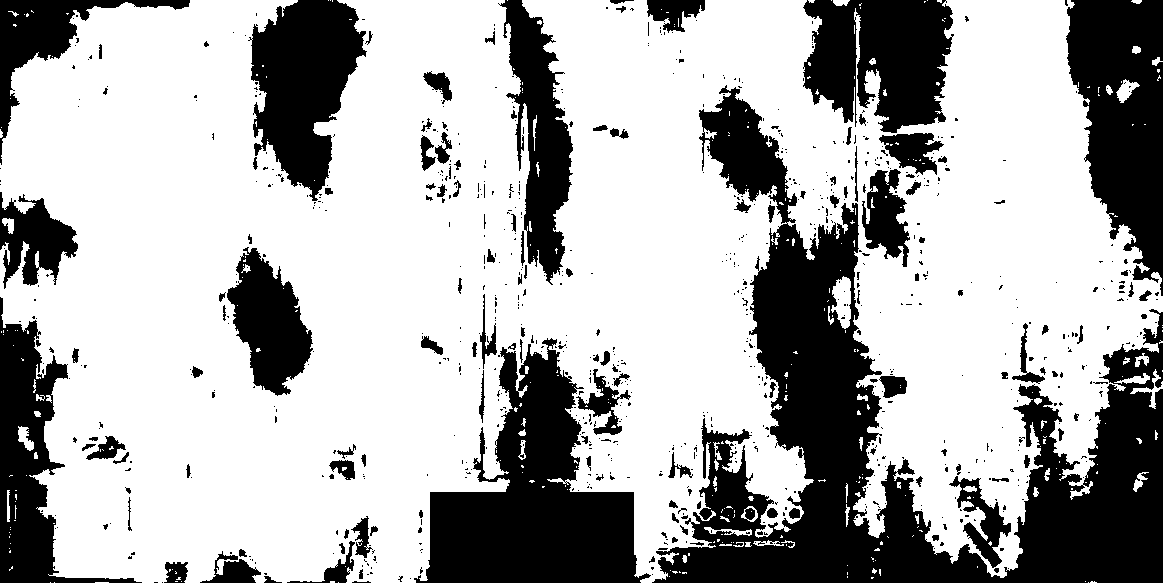}}
        \label{fig:liq_disp_b}
    }%
    \caption{\blue{Red-colored water was ejected by the nozzle on a PV panel covered with cardboard sheets. (a) shows an RGB image of the surface sprayed with red-colored water. (b) shows the corresponding segmentation into coated and uncoated regions (segmented in white and in black respectively), obtained by applying an HSV color filter to (a). This filter was manually tuned to isolate the red-colored water from the light brown cardboard background. The fraction of white pixels in (b) is used to compute the $\sim70\%$ coating coverage.}}
    \label{fig:liq_disp}
    \vspace{-0.4cm}
\end{figure}

\blue{To evaluate spray coverage performance, an additional outdoor test is conducted using red-colored water to visualize the coating distribution on the PV panel. Despite windier conditions than previous tests, the quadcopter achieved approximately $70\%$ coverage of the panel surface, as shown in Fig.~\ref{fig:liq_disp}.}

Outdoor experiments successfully demonstrate the transition from laboratory to field conditions. Some videos demonstrating our system can be found in the supplementary materials.
Compensation systems are essential for outdoor operation, effectively reducing altitude drift from ground effect and mass changes. While outdoor testing shows increased position errors due to wind disturbances compared to indoor results, the compensation algorithms maintain consistent performance across environments, confirming their robustness for field deployment. Using a dense array of nozzles instead of a single one would improve the coverage, which is impacted by disturbances such as wind. Active wind estimation from IMU data could be another possible approach to disturbance rejection. We leave these additions to future work.

\section{Conclusion and Future Work}

This paper presents an UAV system for autonomous PV panel coating reapplication. The system integrates PV panel corner detection, VIO-based localization, lightweight liquid dispersion, and compensation algorithms for ground effect and mass variation disturbances. Indoor and outdoor experiments validate the system's ability to autonomously perform close-proximity coverage of PV panels. Results show that compensation systems significantly improve altitude tracking performance, with $z$-axis RMSE reduced from $8.9~\rm{cm}$ to $3.0~\rm{cm}$ during outdoor panel coverage.

\blue{Assuming daylight-only operation, a three-minute coating time per panel, and a $\sim40\%$ up time, we estimate that a single UAV could service on the order of $10^{5}$ panels approximately every 8 years, which is the recommended recoating time of PV panels. If using a quadcopter fleet, this characteristic time estimate would be scaled by the inverse of the fleet size}.

Future work could focus on improving detection accuracy through more curated, PV panel-specific training datasets and more robust 3D corner extraction for close-proximity operations. Wind-robust control approaches such as model predictive control could further improve trajectory tracking performance. To scale to larger PV installations, multi-nozzle spray arrays would significantly increase the coverage area achievable in a single flight and allow for less tightly packed sweeping trajectories.

\section*{Acknowledgment}
This work has been partially supported by the Cal-Next Solar Center, \url{cal-next.berkeley.edu/} and the UC Berkeley College of Engineering. The experimental testbed at the HiPeRLab is the result of contributions of many people, a full list of which can be found at \url{hiperlab.berkeley.edu/members/}.

\bibliographystyle{ieeetr}
\balance
\bibliography{refs}

\end{document}